\documentclass[accepted]{uai2023} % after acceptance, for a revised version; also before submission to see how the non-anonymous paper would look like
\usepackage[american]{babel}
\usepackage{natbib} % has a nice set of citation styles and commands
\usepackage{mathtools} % amsmath with fixes and additions
\usepackage{booktabs} % commands to create gOOD-looking tables
\usepackage{tikz} % nice language for creating drawings and diagrams
\usepackage{algorithm}
\usepackage{algorithmic}
\usepackage{hyperref}
\usepackage{amsmath}
\usepackage{amssymb}
\usepackage{mathtools}
\usepackage{amsthm}
\usepackage[textsize=tiny]{todonotes}
\usepackage{multirow}
\usepackage{microtype}
\usepackage{subfigure}
\usepackage{booktabs}
\usepackage{dsfont}
\theoremstyle{plain}
\newtheorem{theorem}{Theorem}[section]

\theoremstyle{definition}
\newtheorem{definition}[theorem]{Definition}

\theoremstyle{remark}

\title{Mitigating Transformer Overconfidence via Lipschitz Regularization}

\author[1,4]{\href{mailto:<wenqian@virginia.edu>?Subject=Your UAI 2023 paper}{Wenqian Ye}{}}
\author[2,4]{\href{mailto:<yunsheng@purdue.edu>?Subject=Your UAI 2023 paper}{Yunsheng Ma}}
\author[3,4]{\href{mailto:<xucao2@illinois.edu>?Subject=Your UAI 2023 paper}{Xu Cao}}
\author[5]{Kun Tang}
\affil[1]{%
    Department of Computer Science, University of Virginia, Charlottesville, VA, USA
}
\affil[2]{%
   College of Engineering, Purdue University, West Lafayette, IN, USA
}
\affil[3]{%
    Department of Computer Science, University of Illinois Urbana-Champaign, Urbana, IL, USA
  }
\affil[4]{%
    AI Lab, Shenzhen Children’s Hospital, Shenzhen, China
  }  
\affil[5]{%
    T Lab, Tencent, Beijing, China
  }  
  
\begin{document}
\maketitle

\begin{abstract}
  Though Transformers have achieved promising results in many computer vision tasks, they tend to be over-confident in predictions, as the standard Dot Product Self-Attention (DPSA) can barely preserve distance for the unbounded input domain. In this work, we fill this gap by proposing a novel Lipschitz Regularized Transformer (LRFormer). Specifically, we present a new similarity function with the distance within Banach Space to ensure the Lipschitzness and also regularize the term by a contractive Lipschitz Bound. The proposed method is analyzed with a theoretical guarantee, providing a rigorous basis for its effectiveness and reliability. Extensive experiments conducted on standard vision benchmarks demonstrate that our method outperforms state-of-the-art single forward pass approaches in prediction, calibration, and uncertainty estimation. 
\end{abstract}

\section{Introduction}

\begin{figure*}[t]%
\centering
\subfigure[Deep Ensemble]{%
\label{fig:ens}%
\includegraphics[width=0.23\textwidth]{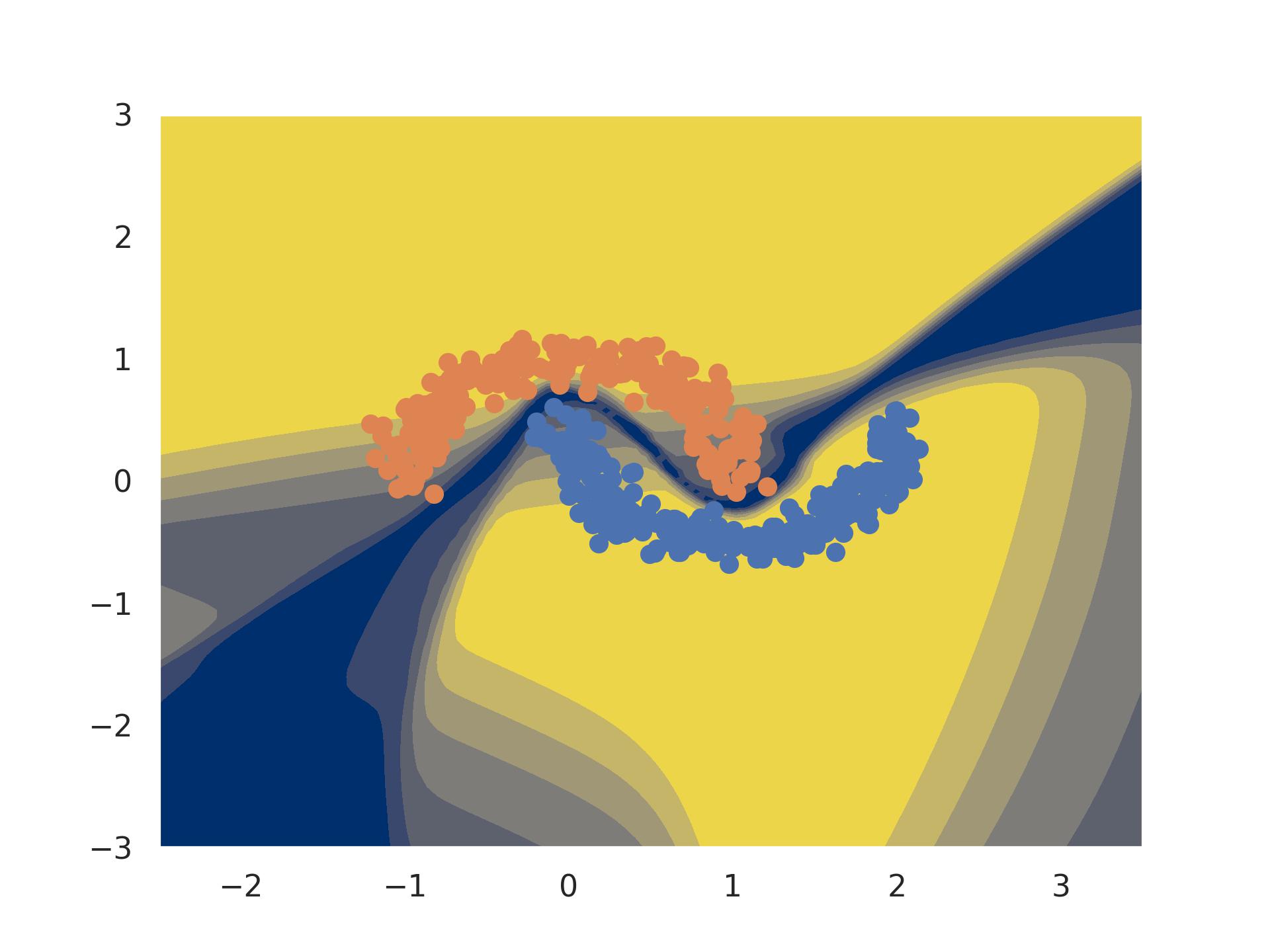}
\includegraphics[width=0.23\textwidth]{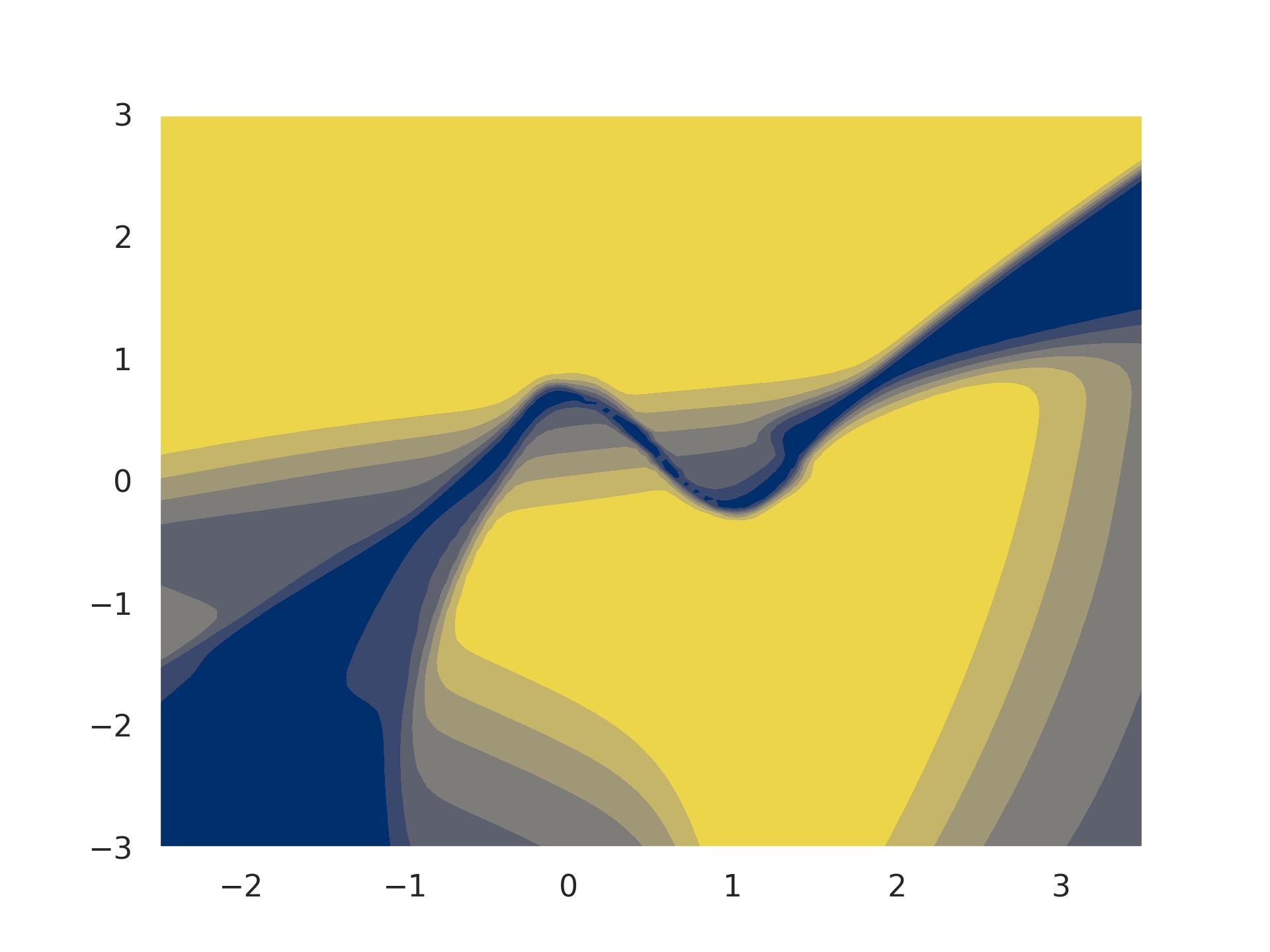}
}%
\subfigure[DUE]{%
\label{fig:due}%
\includegraphics[width=0.23\textwidth]{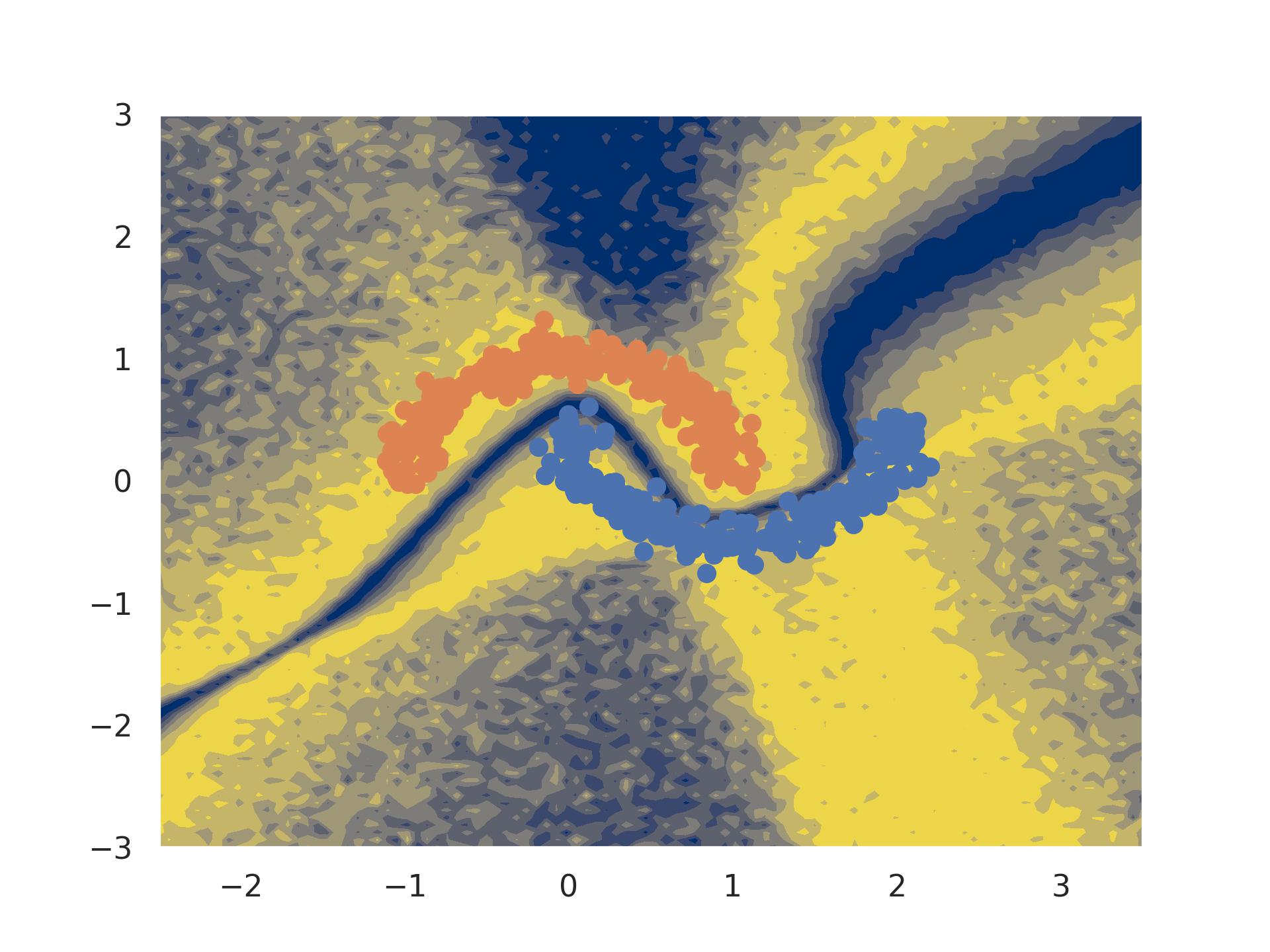}
\includegraphics[width=0.23\textwidth]{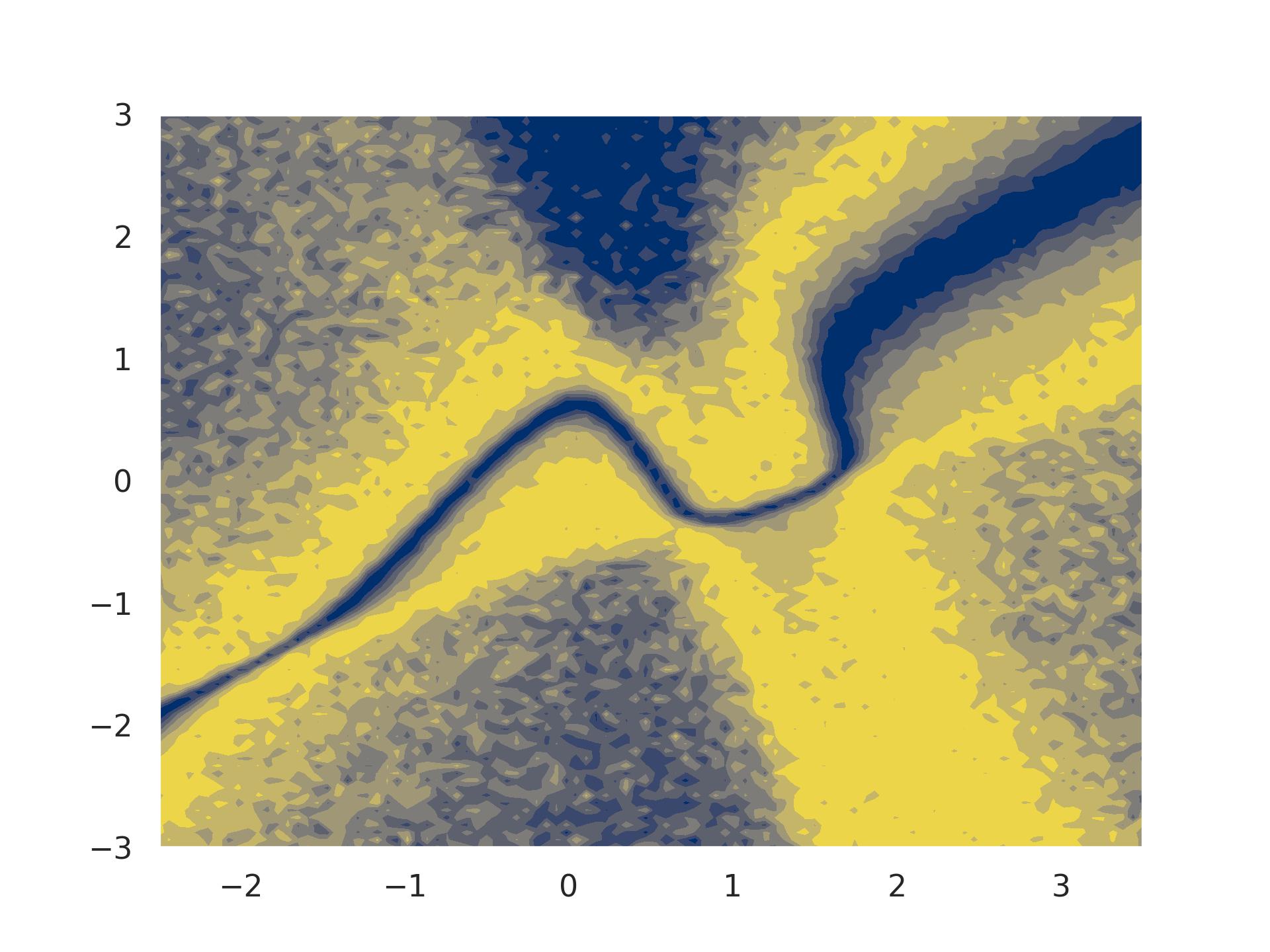}
}\\
\subfigure[SNGP]{%
\label{fig:sngp}%
\includegraphics[width=0.23\textwidth]{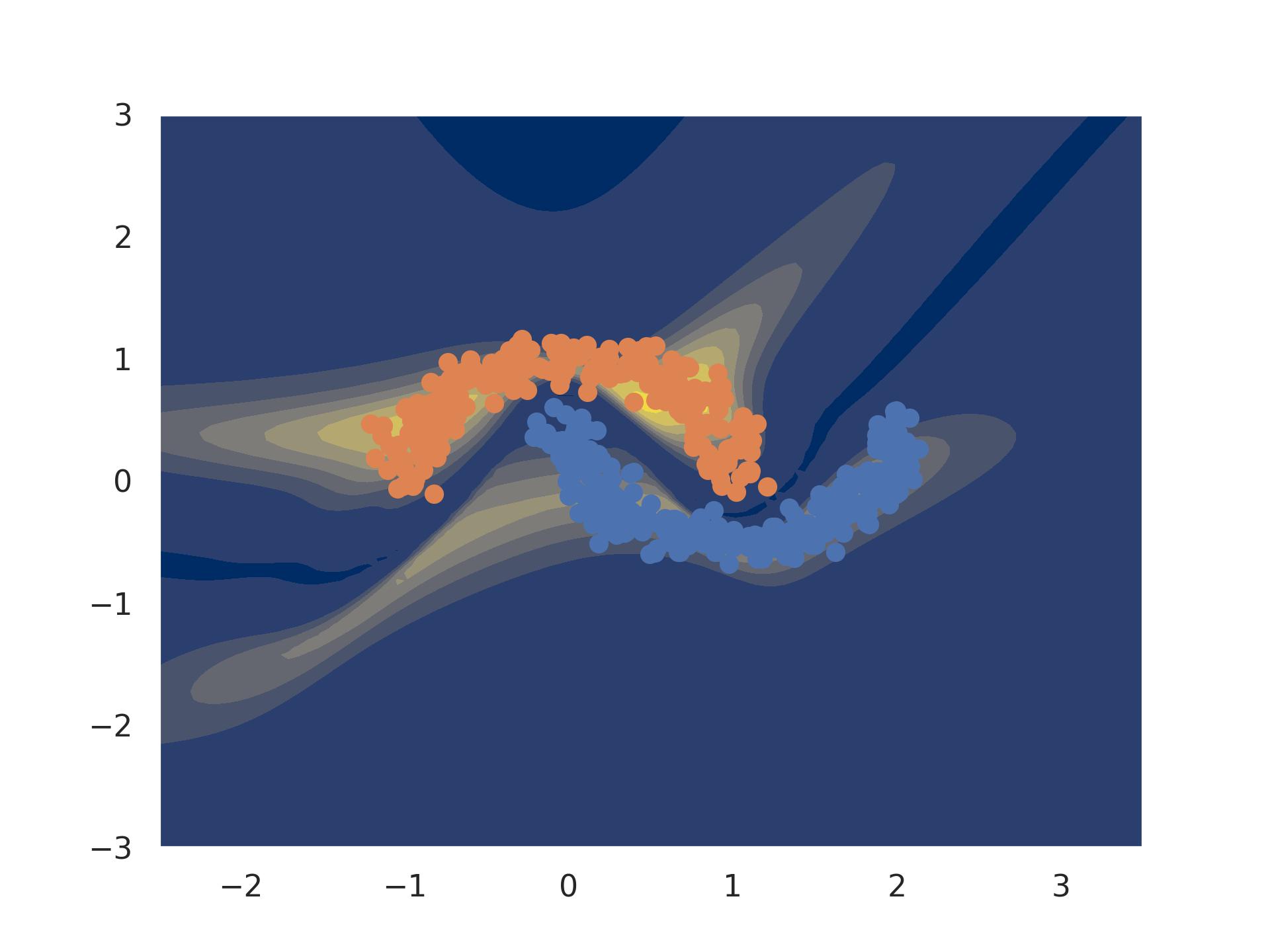}
\includegraphics[width=0.23\textwidth]{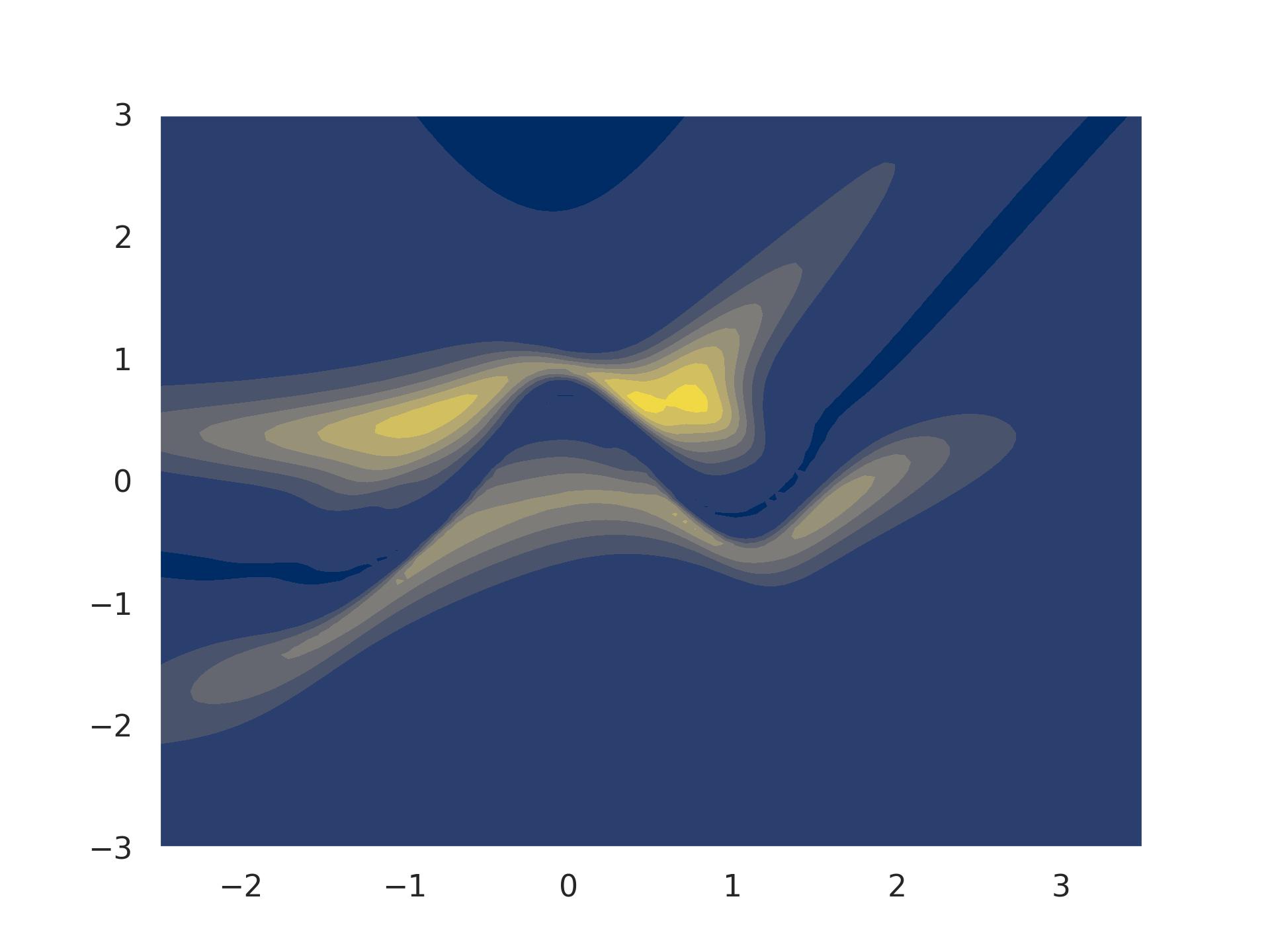}
}%
\subfigure[\textbf{LRFormer}]{%
\label{fig:LRFormer}%
\includegraphics[width=0.23\textwidth]{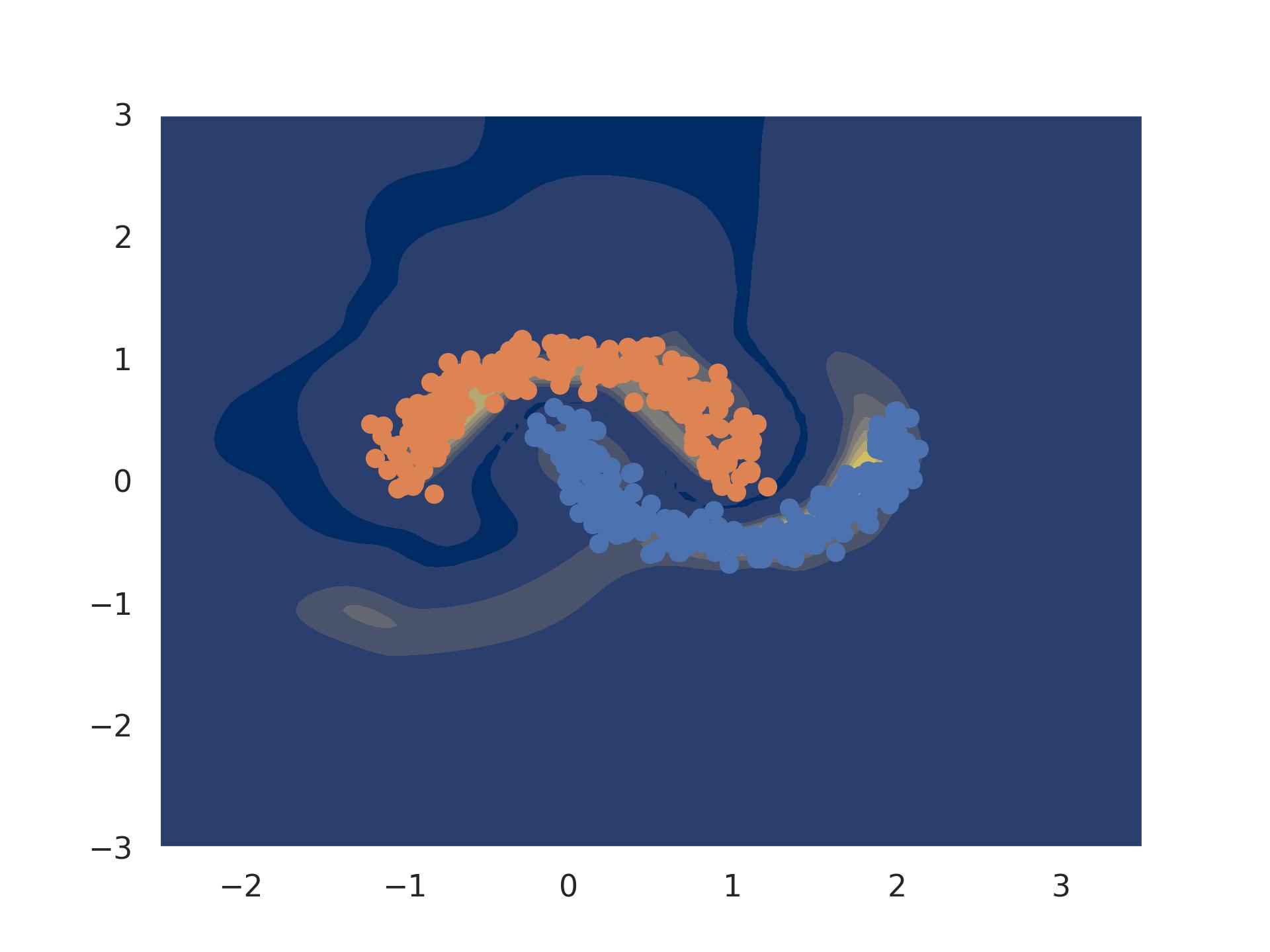}
\includegraphics[width=0.23\textwidth]{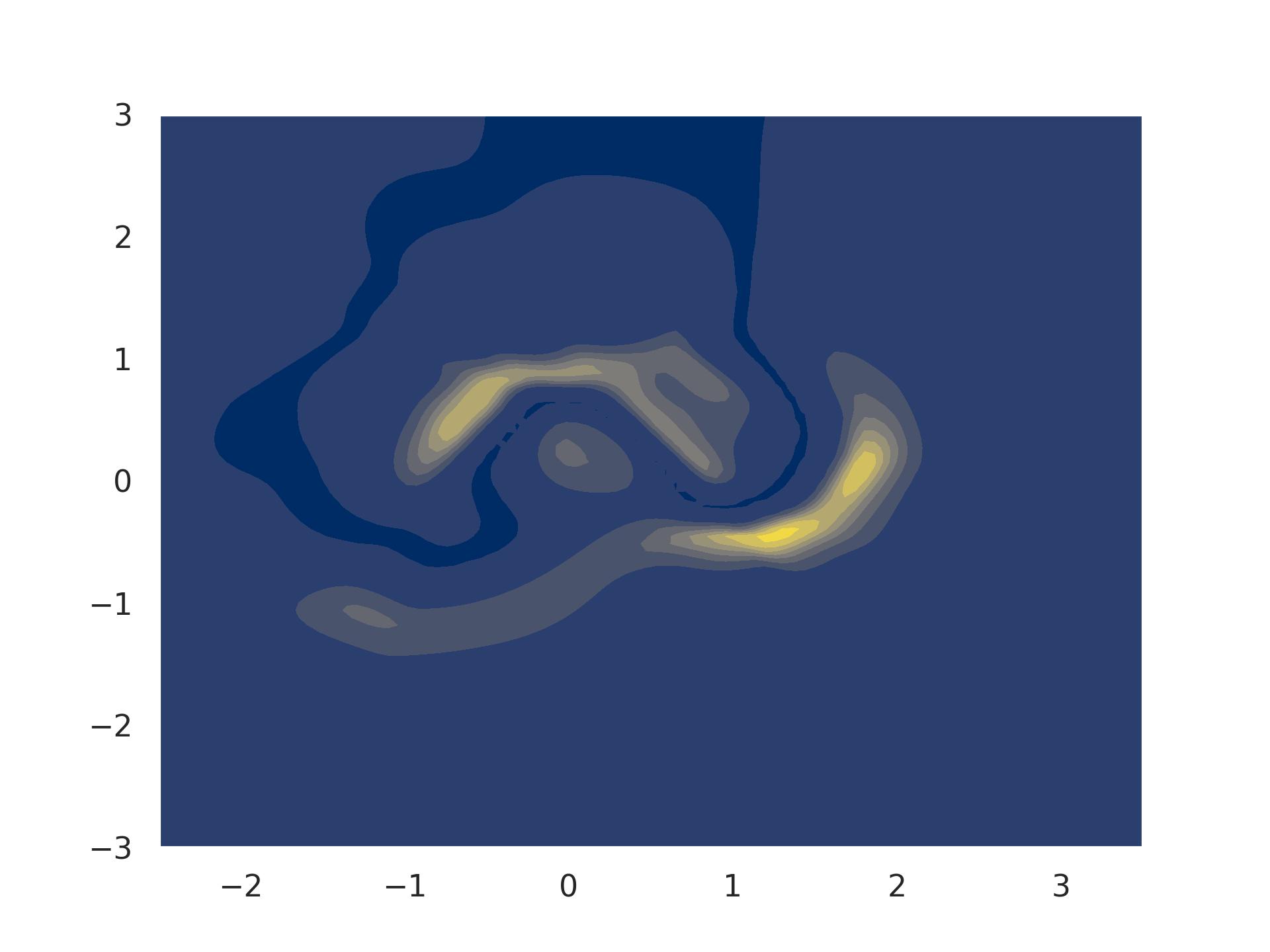}
}%
\caption{Uncertainty heat map of LRFormer and baseline approaches on the two moons 2D classification benchmark. Orange and blue points are positive and negative training samples respectively. Background color visualizes the predictive uncertainty of each model, where \textcolor{yellow}{yellow} stands for confidence and \textcolor{blue}{blue} indicates uncertainty. The proposed LRFormer (Figure \ref{fig:LRFormer}) achieves the closest to ideal uncertainty quantification on this benchmark. Detail refer to Section \ref{sec:two_moons}.}
\label{fig:two_moon}
\end{figure*}

Deep learning (DL) has achieved remarkable performance, making it widely employed in various inference and decision-making systems. However, DL models still make mistakes, making trust and safety an increasingly important topic~\citep{Amodei2016ConcretePI, Jiang2018ToTO}, especially in critical applications like self-driving cars~\citep{Huang2020AutonomousDW} and medical diagnosis~\citep{Esteva2017DermatologistlevelCO}. One solution to this problem is for models to not only achieve high accuracy but also refrain from making overly confident predictions.

Transformer~\citep{Vaswani2017AttentionIA} and its variants, such as BERT~\citep{Devlin2019BERTPO}, have made significant advances in Natural Language Processing (NLP). Similarly, Vision Transformers (ViT)~\citep{Dosovitskiy2021AnII} and its variants~\citep{liu2021swin,li_mvitv2_2022} have recently achieved state-of-the-art performance on a variety of computer vision tasks~\citep{yang_bevformer_2023,hu_planning-oriented_2023,ma_vit-dd_2023,ma_m2dar_2023,xu_cobevt_2022}. Despite this, their propensity for overconfident predictions is cause for concern, especially as they become one of the foundation architectures of deep learning. To address this issue, we investigate the under-explored problem of overconfidence issue in Transformers, which can aid subsequent tasks in the construction of reliable models.

Overconfidence is a common problem in many machine learning models for both in- and out-of-distribution inputs, including deep neural networks~\citep{wei2022mitigating, kristiadi2020being}. When a model is overconfident, it tends to make highly confident predictions even when it is uncertain about the truth of a given input. This can lead to poor performance and inaccurate results, especially in real-world settings where uncertainty is prevalent. Uncertainty estimation is a promising approach for addressing the issue of overconfidence in machine learning models. By estimating uncertainty, a model can make more informed predictions and provide a measure of confidence for each prediction. This can help to improve the robustness and reliability of the model and enable it to perform more effectively in a variety of applications, including decision-making and risk assessment.

Previous techniques for estimating the model's predictive uncertainty include Bayesian deep learning \citep{Wilson2020BayesianDL,blundell2015weight} and ensemble techniques\citep{Lakshminarayanan2017SimpleAS, gal2016dropout}. However, multiple forward passes at the test time are required by most of these methods. In other words, these methods suffer heavy memory and computation cost, which limits their adoption in real-world applications.

Recently, uncertainty quantification via single forward-pass neural networks, which has similar latency as a single deterministic network, has received lots of attention \citep{Liu2020SimpleAP,Amersfoort2021OnFC,Gillioz2020OverviewOT}. SNGP \citep{Liu2020SimpleAP} replaces the dense output layer with a Gaussian Process (GP) layer and applies Spectral Normalization (SN) \citep{Miyato2018SpectralNF} to the hidden residual layers. DUE \citep{Amersfoort2021OnFC} builds upon GPDNN \citep{Bradshaw2017AdversarialEU} and introduces additional constraints to the feature extractor in the form of residual connections in combination with SN ~\citep{Miyato2018SpectralNF}. These methods perform well on uncertainty estimation. However, they only focus on bounding the Lipschitz constants of certain CNN modules \textit{i.e.}, convolution and batch normalization \citep{Ioffe2015BatchNA} layers. Moreover, according to \citet{lee2021vitgan}, Transformer blocks are very sensitive to the magnitude of the Lipschitz constant, and if SN is employed in self-attention modules, training will progress exceptionally slowly. Although some newest proposed Transformer architectures have been proven to be Lipschitz continuous~\citep{qilipsformer,Kim2021TheLC,dasoulas2021lipschitz,xucertifiably, gupta2023certvit}, they still have not solved the overconfidence problem of Transformer.

To address the issues above, we contribute as follows:
\begin{itemize}
    \item We propose a novel regularization technique, termed Lipschitz Regularized Self-Attention (LRSA), that addresses distance awareness in both Lipschitzness and Contraction. LRSA replaces the dot product similarity with the distance within Banach Space and normalizes the term by a theoretical bound of the Lipschitz constant. Furthermore, we provide a theoretical analysis of how our method achieves these properties.
    \item We develop the LRSA based Transformer called LRFormer\footnote{\url{https://github.com/SZCHAI/LRFormer}}, which integrates distance-preserving hidden mappings in transformer blocks via LRSA and utilizes an optional Gaussian Process (GP) distance-aware output layer for high-quality uncertainty estimation. 
    \item We conduct extensive experiments on widely used OOD benchmarks, including CIFAR-10/-100 versus SVHN and CIFAR-10/-100 versus CIFAR-100/-10. Compared to state-of-the-art approaches, our experimental results demonstrate that the proposed LRFormer model is superior in terms of prediction, calibration, and uncertainty estimation, with minimal time complexity penalty.
\end{itemize}

\section{Problem Statement}

In the supervised multi-class classification setting, assume a data sample $(\boldsymbol{x}, y) \in \mathcal{X} \times \mathcal{Y}$ is sampled from an unknown distribution, where $\mathcal{Y}=\{1, \ldots, K\}$ denote the label space with $K$ classes and $\mathcal{X}=\mathbb{R}^d$ denote the feature space. A learned classifier $f^\theta$: $\mathcal{X} \rightarrow \Delta^K$ can produce a probability distribution for $\boldsymbol{x}$ on $K$ classes, where $\Delta^K$ is the $K-1$ dimensional unit simplex. In this context, we introduce a former definition of overconfidence for a general classifier.

\begin{definition}[Overconfidence]
\label{def:overconfidence}
Assume $f^\theta$ as a composition of a non-probabilistic $K$-way classifier $\boldsymbol{h}^\theta$ and a softmax function $\sigma$, i.e. $\boldsymbol{f}^\theta=\boldsymbol{h}^\theta \circ \sigma$. Given a test data sample $\boldsymbol{x}, \boldsymbol{f}^\theta$ provides its probability of assigning it to label $i$ as $\frac{\exp \left(h_i^\theta(\boldsymbol{x})\right)}{\sum_{k=1}^K \exp \left(h_k^\theta(\boldsymbol{x})\right)}$, where $h_i^\theta(\boldsymbol{x})$ denotes the $i$-th element of the logit vector produced by $\boldsymbol{h}^\theta$. Then, $\hat{y}:=\arg \max _i f_i^\theta(\boldsymbol{x})$ can be returned as the predicted label and $\hat{p}:=\max _{j \neq y} f_j^\theta(\boldsymbol{x})$ can be treated as the confidence score. Overconfidence appears when the prediction is wrong with high probability.
\end{definition}

\begin{figure}
    \centering
    \includegraphics[width=0.92\linewidth]{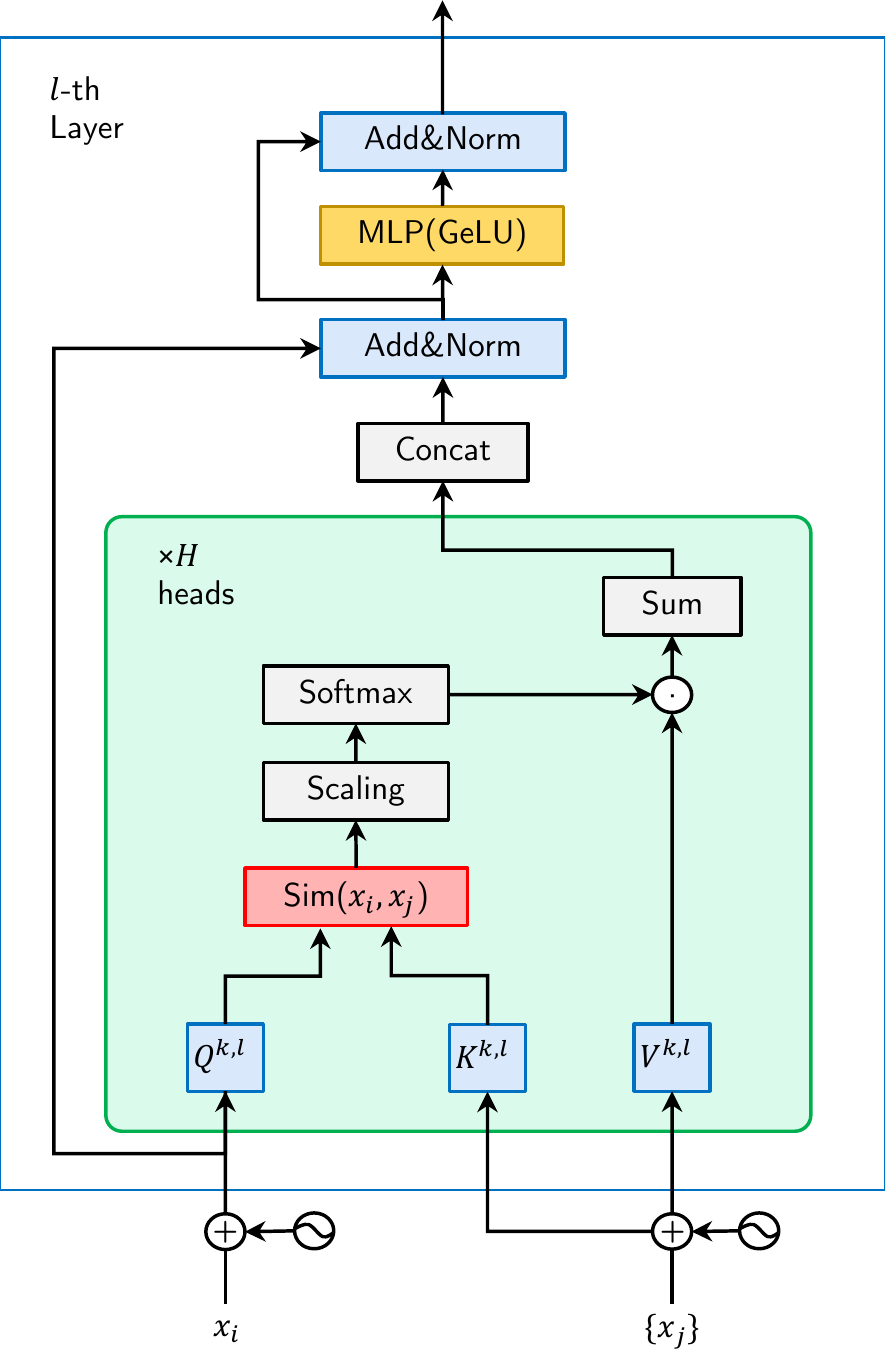}
    \caption{By breaking down a Transformer Layer into its fundamental components, including GeLU Activation, MLP Layer, LayerNorm, and the Attention Module, we gain a comprehensive understanding of their individual contributions to the analysis of Lipschitzness in Transformers. Detailed analysis can be found in Section \ref{sec:analysis}.}
    \label{fig:lrsa}
\end{figure}

Based on Definition \ref{def:overconfidence}, the logit vector can be decomposed into two components: $\boldsymbol{h}(\boldsymbol{x}) = ||\boldsymbol{h}(\boldsymbol{x})|| \cdot \hat{\boldsymbol{h}(\boldsymbol{x})}$, where $||\boldsymbol{h}(\boldsymbol{x})||$ is the L2-norm of the logit vector and $\hat{\boldsymbol{h}(\boldsymbol{x})}$ is the unit vector in the same direction as $\boldsymbol{h}(\boldsymbol{x})$. These two terms represent the magnitude and direction of the logit vector, respectively. It is evident that if $\arg \max_k (h_k)=c$, then $\arg \max_k (\gamma \boldsymbol{h}_k)=c$ always holds for any given constant value $\gamma>1$. This indicates that the magnitude of the logit vector does not affect the predicted class $c$. Additionally, for any given scalar $\gamma>1$, if $c=\arg \max_k (h_k)$, then $\sigma_c(\gamma \boldsymbol{h}) \geq \sigma_c(\boldsymbol{h})$. From the above claims, we observe that increasing the magnitude $||\boldsymbol{h}(\boldsymbol{x})||$ will lead to a higher softmax confidence score while leaving the final prediction unchanged.

During optimization, the cross-entropy loss is given as:

$$
\mathcal{L}_{\mathrm{CE}}(\boldsymbol{h}(\boldsymbol{x} ; \theta), y)=-\log p(y \mid \boldsymbol{x})=-\log \frac{e^{\|\boldsymbol{h}\| \cdot \hat{h}_y}}{\sum_{i=1}^k e^{\|\boldsymbol{h}\| \cdot \hat{h}_i}}
$$

While the direction $\hat{\boldsymbol{h}(\boldsymbol{x})}$ remains constant, increasing the magnitude will lead to a smaller $p(y \mid \boldsymbol{x})$. In the standard Transformer, optimization on the training loss leads to an increase in the magnitude of the network output to produce a higher softmax confidence score, resulting in a smaller loss.

Lipschitzness can help a classifier tackle the overconfidence issue by limiting the amount of change in the classifier's output when the input is perturbed slightly. When a classifier is overconfident, it tends to assign high confidence to incorrect predictions, which can result in poor performance on unseen data. However, if the classifier is Lipschitz continuous, then the amount of change in its output is limited when the input is perturbed, which can prevent the classifier from making overly confident predictions on Out-of-Distribution (OOD) samples.

\begin{definition}[Lipschitz Continuity]
Lipschitz constant of a function $f$ is an upper bound on the ratio between the output and the input variations of a function $f$. If $L \in[0,+\infty)$ is such that, for every input $x \in \mathbb{R}^{d}$ and perturbation $\Delta x \in \mathbb{R}^{d}$,
\begin{equation}
    \|f(x+\Delta x) - f(x)\|_p \leqslant L\|\Delta x\|_p
\end{equation}

then $L$ is a Lipschitz constant of $f$. $\|\cdot\|_p$ denotes the p-norm. If $X$ is defined as the $\epsilon$-ball at point $x$, i.e., $\mathcal{X'}=\left\{x^{\prime} \mid\left\|x-x^{\prime}\right\| \leqslant \epsilon\right\}$, then $L$ is the local Lipschitz constant of $f$ at $x$. 

Furthermore, the Lipschitz condition can be extended to the Bi-Lipschitz condition. Given any two input samples $x_1, x_2$ and a non-probabilistic K-way classifier $\boldsymbol{h}^\theta$, the Bi-Lipschitz condition can be defined as:

\begin{equation}
\label{eq:att}
L_{1} \|x_1 - x_2\| \leq\|\boldsymbol{h}^\theta(x_1)-\boldsymbol{h}^\theta(x_2)\| 
\leq L_{2} \|x_1 - x_2\|
\end{equation}
where $\|.\|$ is a semantically meaningful distance for the data manifold for positive and bounded Lipschitz constants $0<L_{1}<1<L_{2}$. These bounds $L_{1}$, $L_{2}$ represent \textit{sensitivity} and \textit{smoothness} conditions which prevents the hidden representations $\boldsymbol{h}^\theta(x)$ from being unnecessarily
invariant to the semantically meaningful changes in the input manifold or being overly sensitive to the semantically meaningless perturbations respectively.
\end{definition}

Dot-Product Self-Attention is a fundamental building block of the Transformer model. It enables the model to focus on the most relevant parts of the input sequence by weighing the contribution of each input vector to the output, as given by the equation $\operatorname{Attention}(X)= S(X) \cdot V(X) = \operatorname{softmax}\left(\frac{Q\cdot K^\top}{\sqrt{d_k}}\right)\cdot V(X)$. This mechanism can be further generalized by incorporating a similarity function that measures the relevance between input vectors ~\citep{katharopoulos2020transformers}.  While several similarity functions, such as the cosine similarity \citep{qilipsformer} or the scaled dot product, have been used in the original formulation, they may not be optimal for all scenarios. Therefore, in this paper, we explore the use of a Lipschitz similarity function to mitigate overconfidence issues in the Transformer. We present a method for constructing a suitable Lipschitz similarity function and demonstrate its effectiveness in improving the robustness and accuracy of the model.

\begin{equation}
    \operatorname{Attention}(\mathbf{x}_{i}, \mathbf{x}_{j})= \operatorname{softmax}\left(\frac{\operatorname{sim}(\mathbf{x}_{i}, \mathbf{x}_{j})}{\sqrt{d_k}}\right)\cdot V
\end{equation}

Our method aims to maintain a reasonable Lipschitz constant at the block level to address the issue of overconfidence in neural networks. While other methods, such as Bayesian approaches and label smoothing~\citep{muller2019does}, have been proposed to tackle overconfidence, our method incorporates block-wise control. By constraining the Lipschitz constant at each block in the Transformer, we can limit the growth of the magnitude of the network output and reduce overconfidence. This block-wise design also allows our method to be easily integrated into various Transformer-based architectures, including those that have undergone large-scale pretraining. As a result, our method can be scaled up to handle a wide range of tasks and datasets.

\section{Our Method}

\subsection{Notations and Setup}
\begin{itemize}
    \item $S^{(i)}:=\operatorname{diag}\left(S_{i:}\right)-S_{i:}^{\top}, S_{i:} \in \mathbb{R}^{N \times N} .$
    \item Binary Matrix with one in the $(i, j)$ the entry and zeros elsewhere: $E_{i j} \in \mathbb{R}^{N \times N}$
    \item Kronecker delta: $\delta_{i j} \in\{0,1\}$ 
    \item $(\infty, 2)$-norm: $\|M\|_{(\infty, 2)}=\max _{i}\left(\sum_{j} M_{i j}^{2}\right)^{1 / 2}$
    \item Frobenius norm: $\|M\|_{F}=\left(\sum_{i, j} M_{i j}^{2}\right)^{1 / 2}$
    \item Lipschitz constant $L_{\mathbb{X}, \mathbb{Y}}(f)$: for a function $f: \mathbb{X} \rightarrow \mathbb{Y}$, $L_{\mathbb{X}, \mathbb{Y}}(f)=\sup _{X \in \mathbb{X}} \|\frac{\partial f(X)}{\partial X} \|_{\mathbb{X}, \mathbb{Y}}$
\end{itemize}

\subsection{Lipschitz Regularization on Self Attention}

\citet{Kim2021TheLC} proved that the Scaled Dot-Product Self-Attention does not satisfy the \textit{bi-Lipschitz condition}. To extend the generality of self-attention with high-quality uncertainty estimation, we propose a new regularization method Lipschitz Regularized on Self Attention (LRSA) by replacing the self-attention function with a contractive Bi-Lipschitz expression without losing the original ability of representation. We will explicitly discuss separate aspects to see how to achieve both Lipschitzness and Contraction in our method.

\subsubsection{Lipschitzness}
Given that \textit{Dot-Product Self-Attention is not Lipschitz}, suppose there exists such mapping $f(X)$, $X \in \mathbb{R}^N$:
$$
         f(X) = S\cdot X = \operatorname{softmax}(a X\cdot X^\top) \cdot X = \\ \left[\begin{array}{c}
            f_{1}(X) \\
            \vdots \\
            f_{N}(X)
            \end{array}\right] 
$$
Its Jacobian Matrix is $J_f = [J_{ij}]_{N\times N}$, each entry can be written as:
 $$J_{i j}=a X^{\top} S^{(i)}\left[E_{j i} X+\delta_{i j} X\right]+S_{i j} I \in \mathbb{R}^{N \times 1}$$
 
Thus for $i=j$: 
\begin{equation}
J_{i i}=a X^{\top} S^{(i)} E_{i i} X+a X^{\top} S^{(i)} X+S_{i i}
\end{equation}
$X^{\top} S^{(i)} X$ is in the form of a variance of a discrete distribution. When $\mathbf{x}_{i}=\mathbf{0}$ for some $i$, some entries of the Jacobian of $f$ grow proportionally to the sample variance of $\mathbf{x}_{\neq i}$.(The softmax probabilities $S_{i:}$ are constant with respect to $\mathbf{x}_{\neq i}$ when $\mathbf{x}_{i}=0$.)  This will lead to an unbounded Jacobian matrix. 

To avoid this pathology, we replace $Q \cdot K^{\top}$ by $\operatorname{sim}(\mathbf{x}_{i}, \mathbf{x}_{j}) = - \|\mathbf{x}_{i}^{\top} Q-\mathbf{x}_{j}^{\top} K\|_{2}^{2}$ in $\operatorname{Attention}(X)$. Here, the new similarity measurement lies in the Banach Space (complete vector space with norm $\|\cdot\|$), which is a more generalized space over Hilbert Space (complete inner product space) \citep{megginson2012introduction}. This modification also gives a strong theoretical guarantee on 
Lipschitzness with easy matrix multiplications during training. 

\subsubsection{Contraction}
Contraction of the Scaled Dot-Product Self-Attention is another crucial issue for achieving well-calibrated uncertainty. Deriving such contraction scalar requires a theoretical lower bound of the Lipschitz constant on the Dot-Product Self-Attention function. A desirable contraction scalar could be non-strict but easy to compute during training.

\begin{theorem}[\citet{dasoulas2021lipschitz}] \label{thm:3}
For $\alpha \ge 0$, if $\tilde{g}$ is Lipschitz and for all $X \in \mathbb{R}_{d \times n}$, and $\tilde{g}$ satisfy the following conditions:
\begin{enumerate}
    \item $\|\tilde{g}(X)\|_{\infty} \leqslant \alpha c(X)$,
    \item $\|X^{\top} \|_{(\infty, 2)} \| \frac{\partial \tilde{g}(X)}{\partial X} \|_{F,(2, \infty)} \leqslant \alpha c(X)$,
    \item $\|X^{\top} \|_{(\infty, 2)} \|\frac{\partial c(X)}{\partial X} \|_{F, 1}\|\tilde{g}(X)\|_{(2, \infty)} \leqslant \alpha c(X)^{2}$,
\end{enumerate}
where $c$ is a scalar function $c: \mathbb{R}^{d \times n} \rightarrow \mathbb{R}_{+}$. Then $g(X)$ is Lipschitz:

\begin{equation} 
    \begin{aligned}
        g(X) =  \frac{\alpha \tilde{g}(X)}{\max \left\{\|\tilde{g}(X)\|_{(2, \infty)},\left\|X^{\top}\right\|_{(\infty, 2)} L_{F,(2, \infty)}(\tilde{g})\right\}}
    \end{aligned}
\end{equation}
\end{theorem}

Inspired by \ref{thm:3}, we introduce a proper regularization scalar function with a Scalar Factor $\alpha$ by replacing $\tilde{g}(X)$ with $Q \cdot K^{\top}$:
\begin{equation} \label{eq:5}
   c(X) = \frac{\alpha}{\|Q\|_{F} \cdot \|X^{\top} \|_{(\infty, 2)}}
\end{equation}

Here, we assign it as a hyperparameter in control of the corresponding Lipschitz constants for proper contraction of the attention block. Small alpha results in a loss of information while a large alpha causes the model tending to be non-Lipschitz.

\subsubsection{Summary}
Here is the formal definition of the similarity function:
\begin{equation}
S_{i j} := b \cdot c(X) = -\frac{\alpha \|\mathbf{x}_{i}^{\top} W_Q-\mathbf{x}_{j}^{\top} W_K\|_{2}^{2}}{\|Q\|_{F} \cdot \|X^{\top} \|_{(\infty, 2)}}
\end{equation}

 This pair-wise operation can alternatively be implemented as a matrix version for improved computational efficiency:

\begin{equation} \label{LRSA}
  S(X) = \operatorname{softmax}(-\alpha \cdot \frac{\|Q\|_{\text {row }}^{2} -2 Q K^{\top}+ \|K\|_{\text {col }}^{2^{\top}}}{\|Q\|_{F} \cdot \|X^{\top} \|_{(\infty, 2)}})
\end{equation}
LRSA Attention can be represented by the expression $\operatorname{LRSA}(X) = S(X) \cdot V(X)$, where $S(X)$ denotes the similarity scores and $V(X)$ represents the value embeddings. In the following section, we define the Lipschitz Constant of $S(X)$ as $L_{\operatorname{LRSA}}$. From Supplementary Material, we can conclude that $L_{\operatorname{LRSA}}$ is bounded by $\frac{6 \alpha}{\left\Vert X \right\Vert _F} \cdot \frac { (\left\Vert W_Q\right\Vert_2 + \left\Vert W_K \right\Vert_2) }{\left\Vert W_Q \right\Vert _F }^2$.

\subsection{Bottom-Up Analysis on LRFormer}
\label{sec:analysis}

\noindent \textbf{Analysis on Lipschitzness of GeLU Activation}

\begin{figure}
    \centering
    \includegraphics[width=0.9\linewidth]{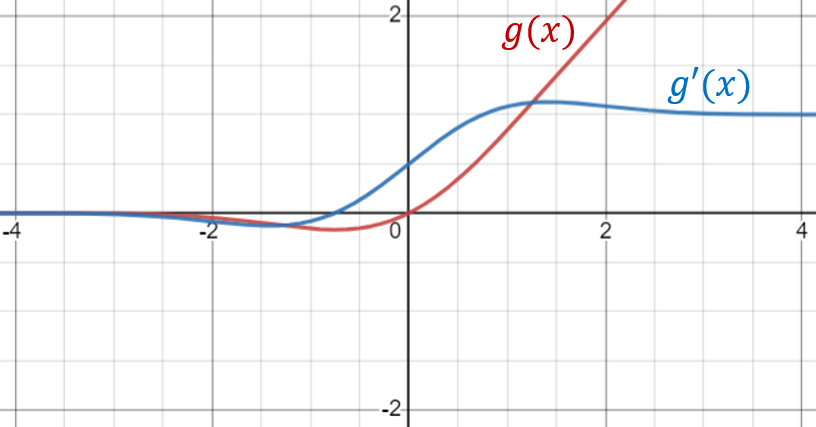}
    \caption{GeLU $g(x)$ and the derivative of GeLU $g'(x)$.}
    \label{fig:gelu}
\end{figure}

GeLU~\citep{Hendrycks2016GaussianEL} is the most commonly used activation function in Transformers, especially the GPT series of models and Vision Transformers. GeLU's activation function form is $\operatorname{GeLU}(x)=x\Phi(x)$, where $\Phi(x)$ is the standard Gaussian cumulative distribution function. The derivative of GeLU is given as:

\begin{equation}
    \operatorname{GeLU}'(x) = \frac{xe^{-\frac{x^2}{2}}}{\sqrt{2\pi}} + \frac{\operatorname{erf}(\frac{x}{\sqrt{2}})}{2} + \frac{1}{2}
\end{equation}

,where $\operatorname{erf}(x)=\frac{2}{\sqrt{\pi}}\int_{0}^{x}e^{-t^2 dt}$. The second derivative of GeLU is:

\begin{equation}
    \operatorname{GeLU}''(x) = \frac{1}{\sqrt{2\pi}} ( (1 + \sqrt{2}) \cdot e^{-\frac{x^2}{2}}-x^2e^{-\frac{x^2}{2}})
\end{equation}

Compared with ReLU, GeLU is differentiable at zero and keeps the Lipschitz continuous. The Lipschitz constant of GeLU is the value of $x$ from $\max (\operatorname{GeLU}'(x))$. Let $\operatorname{GeLU}''(x)=0$, we can verify $\max (\operatorname{GeLU}'(x))\approx1.129$.

\noindent \textbf{Analysis on Lipschitzness of LayerNorm}

The raw LayerNorm operation ($\text{LN}(\mathbf{x}) = \frac{\mathbf{x}-\mu(\mathbf{x})}{\sqrt{\sigma^2(\mathbf{x})}} * \boldsymbol\gamma + \boldsymbol\beta$)~\citep{layernorm} is not Lipschitz continuous because the ill-defined input with zero variance will lead to a Jacobian matrix filled with infinity.

However, the LayerNorm operation can be changed to a Lipschitz continuous form, which is the LayerNorm used in our models. The form can be expressed as:
\begin{equation}
    \begin{aligned}
    \text{LayerNorm}(\mathbf{x}) &= \frac{\mathbf{x}-\mu(\mathbf{x})}{\sqrt{\sigma^2(\mathbf{x}) + \epsilon}} * \boldsymbol\gamma + \boldsymbol\beta 
\end{aligned}
\end{equation}

where $\mathbf{x}, \boldsymbol\beta, \boldsymbol\gamma \in \mathbb{R}^N$, $\mu(\mathbf{x}) = \frac{1}{N} \sum_{i=1}^N x_i$, $\sigma^2(\mathbf{x}) = \frac{1}{N}\sum_{i=1}^N (x_i - \mu(\mathbf{x}))^2$. 

From Supplementary Material, we can conclude that LayerNorm is Lipschitz with the constant $\eta = \epsilon^{-\frac{1}{2}} \max_i |\gamma_i| N$.

\noindent \textbf{Analysis on Lipschitzness of MLP Layer}

In each Transformer block, the Attention Module is typically followed by the Multi-Layer Perceptron (MLP) Layer. The MLP layer consists of two Fully Connected (FC) Layers, a Dropout Layer, and a GeLU activation function. Since the Dropout Layer has no impact on the Lipschitz constant, it can be simplified as:
\begin{equation}
    \operatorname{MLP}(x) = \operatorname{FC}_2 \circ \operatorname{GeLU} \circ \operatorname{FC}_1 (x)
\end{equation}
The upper bound on the Lipschitz constant of a fully-connected layer can be derived by analyzing the effect of each layer independently and considering a product of the resulting spectral norms $\sigma(\cdot)$~\citep{Miyato2018SpectralNF}. 
\begin{equation}
    L_{\operatorname{MLP}} = 1.129 \cdot \sigma(W_1) \cdot \sigma(W_2)
\end{equation}

\noindent \textbf{Lipschitz Constant of LRFormer Layer}

In the LRFormer, we analyze the LRFormer Layer to demonstrate our method follows the Lipschitzness. Each LRFormer Layer can be expressed as:

\begin{equation}
     \begin{aligned}
    \operatorname{LRFormer}_{i}(x) &= \operatorname{LayerNorm}(\operatorname{LayerNorm}(x + \operatorname{LRSA}(x)) \\
    &+ \operatorname{MLP}(\operatorname{LayerNorm}(x + \operatorname{LRSA}(x))))
    \end{aligned} 
\end{equation}

Combined with the previous analysis, we conclude the following Lipschitz bound of a single LRFormer layer:
\begin{equation}
\begin{aligned}
    L_{\operatorname{Layer}} &= \eta_1 \cdot \eta_2 \cdot ((1 + L_{\operatorname{LRSA}}) \\
    &+ 1.129 \cdot \sigma(W_1) \cdot \sigma(W_2) \cdot (1 + L_{\operatorname{LRSA}}))
\end{aligned}
\end{equation}
, where $\eta_1$, $\eta_2$ are the Lipschitz Constant of two LayerNorms.

\section{Experiments}

\begin{table*}[!ht]
\centering
\small
\begin{tabular}{c|ccccc}
    \toprule
    \multirow{2}*{Method}  & \multirow{2}*{Accuracy $(\uparrow)$} & \multirow{2}*{ECE $(\downarrow)$} & \multirow{2}*{NLL $(\downarrow)$} & \multicolumn{2}{c}{OOD AUPR $(\uparrow)$} \\

    & \multicolumn{3}{c}{} & SVHN & CIFAR-100 \\
    \midrule
    Deterministic$^{*}$ &   96.0 $\pm$ 0.01 &  0.023 $\pm$ 0.002 & 0.158 $\pm$ 0.01 & 0.781 $\pm$ 0.01 &  0.835 $\pm$ 0.01 \\
    MC Dropout$^{*}$ &  96.0 $\pm$ 0.01 &  0.021 $\pm$ 0.002 & 0.173 $\pm$ 0.01 & 0.971 $\pm$ 0.01 & 0.832 $\pm$ 0.01 \\
    MCD-GP$^{*}$ &  95.5 $\pm$ 0.02 &  0.024 $\pm$ 0.004 & 0.172 $\pm$ 0.01 & 0.960 $\pm$ 0.01 &  0.863 $\pm$ 0.01 \\
    DNN-SN$^{*}$ &  96.0 $\pm$ 0.01 &  0.025 $\pm$ 0.004 & 0.171 $\pm$ 0.01 & 0.974 $\pm$ 0.01 & 0.859 $\pm$ 0.01 \\
    DNN-GP$^{*}$ &  95.9 $\pm$ 0.02 &  0.029 $\pm$ 0.002 & 0.221 $\pm$ 0.02 & 0.976 $\pm$ 0.01 & 0.887 $\pm$ 0.01 \\
    DUQ$^{*}$ &  94.7 $\pm$ 0.02 &  0.034 $\pm$ 0.002 & 0.239 $\pm$ 0.02 & 0.973 $\pm$ 0.01 & 0.854 $\pm$ 0.01 \\
    DUE$^{*}$ &  95.6 $\pm$ 0.04 &  0.018 $\pm$ 0.002 & 0.187 $\pm$ 0.01 & - & - \\
    SNGP$^{*}$ &  95.9 $\pm$ 0.01 & 0.018 $\pm$ 0.001 & 0.138 $\pm$ 0.01 & 0.990 $\pm$ 0.01 & 0.905 $\pm$ 0.01 \\
    Deep Ensemble$^{*\dagger}$ & 96.6 $\pm$ 0.01 & 0.010 $\pm$ 0.001 & 0.114 $\pm$ 0.01  & 0.964 $\pm$ 0.01 & 0.888 $\pm$ 0.01 \\
    \cmidrule{1-6}
    \textbf{LRFormer} & \textbf{97.2 $\pm$ 0.01} & \textbf{0.012 $\pm$ 0.001} & \textbf{0.100 $\pm$ 0.01} & \textbf{0.993 $\pm$ 0.01} & \textbf{0.911 $\pm$ 0.01}\\
    \bottomrule
\end{tabular}
\caption{Comparison between proposed LRFormer and SOTA methods on CIFAR-10 vs SVHN/CIFAR-100 benchmarks, averaged over 10 seeds. The best method among single-network approaches is highlighted in \textbf{bold}.  $^{*}$Results from the original papers. $^\dagger$ with 10 models.}
\label{tab:sota_cifar10}
\end{table*}

\begin{table*}[!ht]
\centering
\small
\begin{tabular}{c|ccccc}
    \toprule
    \multirow{2}*{Method}  & \multirow{2}*{Accuracy $(\uparrow)$} & \multirow{2}*{ECE $(\downarrow)$} & \multirow{2}*{NLL $(\downarrow)$} & \multicolumn{2}{c}{OOD AUPR $(\uparrow)$} \\
    & \multicolumn{3}{c}{} & SVHN & CIFAR-10 \\
    \midrule
    Deterministic$^{*}$ &  79.8 $\pm$ 0.02 & 0.085 $\pm$ 0.004 & 0.872 $\pm$ 0.01 & 0.882 $\pm$ 0.01 & 0.745 $\pm$ 0.01 \\
    MC Dropout$^{*}$ &  79.6 $\pm$ 0.02 & 0.050 $\pm$ 0.003 & 0.825 $\pm$ 0.01 & 0.832 $\pm$ 0.01 & 0.757 $\pm$ 0.01 \\
    MCD-GP$^{*}$ &  79.5 $\pm$ 0.04 & 0.085 $\pm$ 0.005 & 0.937 $\pm$ 0.01 & 0.873 $\pm$ 0.01 & 0.754 $\pm$ 0.01 \\
    DNN-SN$^{*}$ &  79.9 $\pm$ 0.02 & 0.098 $\pm$ 0.004 & 0.918 $\pm$ 0.01 & 0.879 $\pm$ 0.03 & 0.745 $\pm$ 0.01 \\
    DNN-GP$^{*}$ &  79.2 $\pm$ 0.03 & 0.064 $\pm$ 0.005 & 0.885 $\pm$ 0.01 & 0.876 $\pm$ 0.01 & 0.746 $\pm$ 0.02 \\
    DUQ$^{*}$ &  78.5 $\pm$ 0.03 & 0.119 $\pm$ 0.001 & 0.980 $\pm$ 0.02 & 0.878 $\pm$ 0.01 & 0.732 $\pm$ 0.01 \\
    SNGP$^{*}$ &  79.9 $\pm$ 0.03 & 0.025 $\pm$ 0.012 & 0.847 $\pm$ 0.01 & 0.923 $\pm$ 0.01 & \textbf{0.801 $\pm$ 0.01} \\
    Deep Ensemble$^{*\dagger}$ & 80.2 $\pm$ 0.01 & 0.021 $\pm$ 0.004 & 0.666 $\pm$ 0.02 & 0.888 $\pm$ 0.01 & 0.780 $\pm$ 0.01 \\
    \cmidrule{1-6}
    \textbf{LRFormer} &  \textbf{85.2 $\pm$ 0.03} & \textbf{0.018 $\pm$ 0.005} & \textbf{0.538 $\pm$ 0.01} & \textbf{0.955 $\pm$ 0.01} & 0.777 $\pm$ 0.01 \\    
    \bottomrule
\end{tabular}
\caption{Comparison between proposed LRFormer and the SOTA methods on CIFAR-100 vs SVHN and CIFAR-100 vs CIFAR-10 benchmark, averaged over 10 seeds. The best method among single-network approaches is highlighted in \textbf{bold}. $^{*}$Results from the original papers. $^\dagger$ with 10 models.}
\label{tab:sota_cifar100}
\end{table*}

\begin{table}[!ht]
\centering
\small
\begin{tabular}{c|cccccc}
    \toprule
    Method  & ECE $(\downarrow)$ & NLL $(\downarrow)$ \\
    \midrule
    DP Attention~\citep{Vaswani2017AttentionIA}  & 0.066 & 0.580 \\
    L2 Attention~\citep{Kim2021TheLC} & 0.048 & 0.582  \\
    SCSA~\citep{qilipsformer} & 0.028 &  0.626 \\
    \textbf{LRSA} & \textbf{0.018} & \textbf{0.538} \\
    \bottomrule
\end{tabular}
\caption{Overconfident evaluation comparison among different attention modules. The backbones are the same except for the kernel. We compare the checkpoint when the accuracy achieves $0.85 \pm 0.01$ with CIFAR-100. Note, we select the best hyper-parameter for SCSA ($\nu = 1.0, \tau = 12, \epsilon = 1e-8$)}
\label{tab:ab_attention100}
\end{table}

In this section, we verify the effectiveness of LRFormer in OOD detection with several benchmark datasets. We also design ablation experiments including attention module comparison, searching for a proper scalar factor $\alpha$, and validating the reliability of pretrained models.

\begin{table}[!t]
    \centering
    \small
    \begin{tabular}{c|cc}
    \toprule
        $\alpha$ & Accuracy $(\uparrow)$  & NLL $(\downarrow)$ \\
        \midrule
        1 & 0.5637 $\pm$ 0.01 & 1.7063 $\pm$ 0.02 \\
        100 & \textbf{0.6601} $\pm$ 0.02 & \textbf{1.3712} $\pm$ 0.01 \\
        500 & 0.6404 $\pm$ 0.01 & 1.4041 $\pm$ 0.03 \\
        1000 & 0.6212 $\pm$ 0.02 & 1.4259 $\pm$ 0.01 \\
        \bottomrule
    \end{tabular}
    \caption{Effect of $\alpha$ in LRSA with CIFAR-100}
    \label{tab:alpha}
\end{table}

\subsection{Setup}

\subsubsection{Benchmarks}
We evaluate the performance of the proposed LRFormer model on the OOD benchmark \citep{Miyato2018SpectralNF} using SVHN \citep{Netzer2011ReadingDI} as the OOD dataset for the model trained on CIFAR-10/-100 \citep{krizhevsky2009learning}. OOD data is never seen during training, whereas ID samples are semantically similar to training samples. We also show LRFormer's performance on the Two Moons dataset in Figure \ref{fig:two_moon}.

\subsubsection{Baselines}

Our baselines included the deterministic model and two ensemble models: MC Dropout (with 10 dropout samples) and deep ensembles (with 10 models)~\citep{Lakshminarayanan2017SimpleAS}. All models were trained with a dense output layer and no spectral regularization. Besides, we also compared three single-model approaches: MCD-GP (with 10 samples), DUQ~\citep{van2020uncertainty}, DUE~\citep{Amersfoort2021OnFC}, and SNGP series~\citep{Liu2020SimpleAP} including DNN-SN and DNN-GP. For models that use a GP layer, we kept DL = 1024 and computed the predictive distribution using Monte Carlo averaging with 10 samples. For a fair comparison, we set the backbone with the same parameter magnitude (19.9M parameters for LRFormer, and 36.5M for the SNGP series).

\subsubsection{Evaluation Metrics}
Expected Calibration Error (ECE) \citep{guo2017calibration} quantifies the difference between a model's expected confidence (e.g., the maximum probability score) and its actual accuracy. It achieves this by partitioning all the samples, with $n$ representing the total number of samples, into $M$ equally sized bins based on their confidence scores, then calculating the expected difference between accuracy and the average confidence in each bin. In our task, ECE can indicate the effectiveness of the model in dealing with overconfidence.

In addition to ECE, we employ Negative Log Likelihood (NLL), OOD Area Under the Receiver Operating Characteristic Curve (AUROC), and OOD Area Under the Precision-Recall Curve (AUPR) to evaluate the model's performance in overconfidence and uncertainty estimation ability.
 
\subsubsection{Implementation Details}
In the following experiments, we resize the input image to $224\times 224$ pixels and set the patch size of LRFormer to 16. We employ AdamW~\citep{loshchilov2017decoupled} as the optimizer with a weight decay of 0.05. We use a cosine learning rate scheduler~\citep{loshchilov2016sgdr} with the base learning rate set to $5\times 10^{-5}$. All models are trained for 100 epochs with 10 different random seeds on NVIDIA A100 GPUs.

\begin{table*}[!t]
    \centering
    \small
    \begin{tabular}{l|c|cccc}
    \toprule
       Dataset & Pretained & Accuracy $(\uparrow)$  & NLL $(\downarrow)$ & OOD AUROC $(\uparrow)$ & OOD AUPR $(\uparrow)$  \\
        \midrule
       \multirow{2}{1.5cm}{CIFAR-10} & W/O & 0.8528 $\pm$ 0.01 & 0.4447 $\pm$ 0.02 & 0.8500 $\pm$ 0.01 & 0.9078 $\pm$ 0.02\\
       & W/ & \textbf{0.8616 $\pm$ 0.01} & \textbf{0.4193 $\pm$ 0.01} & \textbf{0.9125 $\pm$ 0.02} & \textbf{0.9499 $\pm$ 0.01}\\
       \midrule
       \multirow{2}{1.68cm}{CIFAR-100} & W/O & 0.6404 $\pm$ 0.01 & 1.4041 $\pm$ 0.03 & 0.8421 $\pm$ 0.01 & 0.9165 $\pm$ 0.01 \\
       & W/ & \textbf{0.6679 $\pm$ 0.01} & \textbf{1.2122 $\pm$ 0.02} & \textbf{0.8689 $\pm$ 0.01} & \textbf{0.9319 $\pm$ 0.01} \\
        \bottomrule
    \end{tabular}
    \caption{Ablation study between with (W/) and without (W/O) pre-trained weights from ImageNet-1K dataset. The best method among single-network approaches is highlighted in \textbf{bold}. $\downarrow$ means lower is better. $\uparrow$ means higher is better.}
    \label{tab:pretrain_tiny}
\end{table*}

\begin{table*}[ht]
\centering
\small
\begin{tabular}{l|c|cccc}
    \toprule
    Dataset & Method & Accuracy $(\uparrow)$ & NLL $(\downarrow)$ & OOD AUROC $(\uparrow)$ & OOD AUPR $(\uparrow)$  \\
    \midrule
    \multirow{5}{1.5cm}{CIFAR-10} 
    & Transformer & \textbf{0.8592 $\pm$ 0.01 } & 0.6972 $\pm$ 0.02 & 0.7552 $\pm$ 0.05 & 0.8521 $\pm$ 0.02 \\
    & DUE + Transformer & 0.8556 $\pm$ 0.03 & 0.5337 $\pm$ 0.02 & 0.8348 $\pm$ 0.04 & 0.8921 $\pm$ 0.01 \\
    & SNGP + Transformer & 0.8542 $\pm$ 0.02 & 0.4933 $\pm$ 0.01 & 0.8275 $\pm$ 0.03 & 0.8960 $\pm$ 0.01 \\
    \cmidrule{2-6}
    & \textbf{LRFormer} & 0.8528 $\pm$ 0.02 & \textbf{0.4447 $\pm$ 0.01} & \textbf{0.8500 $\pm$ 0.05} & \textbf{0.9078 $\pm$ 0.01}  \\
    \midrule
    \multirow{4}{1.68cm}{CIFAR-100} 
    & Transformer & 0.6304 $\pm$ 0.02 & 1.7862 $\pm$ 0.01 & 0.7831 $\pm$ 0.02 & 0.8701 $\pm$ 0.02 \\
    & SNGP + Transformer & 0.6298 $\pm$ 0.03 & 1.5413 $\pm$ 0.01 & 0.8134 $\pm$ 0.01 & 0.8929 $\pm$ 0.01 \\
    \cmidrule{2-6}
    & \textbf{LRFormer} & \textbf{0.6404 $\pm$ 0.02} & \textbf{1.4041 $\pm$ 0.02} & \textbf{0.8421 $\pm$ 0.01} & \textbf{0.9165 $\pm$ 0.01}  \\
    \bottomrule
\end{tabular}
\caption{Ablation study between the proposed LRFormer and existing uncertainty quantification methods with the same training backbone on the CIFAR-10/-100 vs SVHN benchmark. The best method among single-network approaches is highlighted in \textbf{bold}. $\downarrow$ means lower is better. $\uparrow$ means higher is better.}
\label{tab:result_tiny}
\end{table*}

\subsection{Comparison with State-of-the-art models}

Following \citet{Touvron2022ThreeTE}, we adopt an existing training setup, namely the A3 procedure of \citet{wightman2021resnet}. We adjust the learning rate of the A3 procedure when training LRFormer. In our experiments, we set the learning rate to 0.006 for LRFormer when pretraining and 0.004 while finetuning on CIFAR-10/-100. Besides, Different from previous unfair comparison methods \citep{fort2021exploring,xue2022boosting,cao2022deep}, pretrained models from the extra datasets and few-shot outlier exposure settings are not used during training.

To evaluate the model’s OOD detection performance, we adopt the two OOD tasks suggested by SNGP: (1) using SVHN as the OOD dataset for a model trained on CIFAR-10/-100; (2) using CIFAR-100/-10 as the OOD dataset for a model trained on CIFAR-10/-100, respectively. Table~\ref{tab:sota_cifar10} and Table~\ref{tab:sota_cifar100} show the main comparison results. LRFormer outperforms the other single forward pass approaches in all the metrics of CIFAR-10 and most of the metrics of CIFAR-100. Moreover, LRFormer also achieves similar results to Deep Ensemble, which contains 10 models and requires around $10\times$ as much time to execute as LRFormer and other single forward pass approaches.

\subsection{Ablation Study}

\subsubsection{Attention Blocks}
To validate the ability to solve overconfidence issues of Transformers, we compare the LRSA with the scale dot-product attention, L2 attention, and the scaled cosine similarity attention (SCSA) using overconfidence evaluation metrics. In Table\ref{tab:ab_attention100}, we show the best Test ECE and NLL across training for each of the Transformer models. The generalization performance of the best model for each setting of self-attention is similar.

We find that L2 attention, SCSA, and LRSA work well under the Lipschitz guarantee. Meanwhile, LRSA works better than SCSA and L2 attention.

\subsubsection{Hyperparameter Analysis}
The Scalar Factor $\alpha$ in Equation (\ref{eq:5}) controls the scale of the Lipschitz constant of the Transformer blocks. In general, we propose running a grid search for $\alpha\in\left\{...,100, 500, 1000,...\right\}$ to find the highest possible value of $\alpha$ while retaining the predictive performance of LRFormer. In our experiments (Table \ref{tab:pretrain_tiny}), we set scalar factor $\alpha=1000$ in CIFAR-10, and $\alpha=500$ in CIFAR-100. The model's performance is not very sensitive to the parameters in the GP output layer, we follow \citet{Liu2020SimpleAP}'s suggestion and set the number of random features to $1024$, the length-scale for the RBF kernel to $2$, and the $L_2$ regularization to 0. A proper $\alpha$ value, \textit{i.e.} 100, can preserve both the Lipschitzness and Contraction properties in the model. Small alpha will cause loss of information while large alpha will cause the model tending to be non-Lipschitz, leading to degenerate performance.

\subsubsection{Module Comparison}
In this section, we compare LRFormer and other OOD detection methods (using Transformer backbone) under the uncertainty estimation setting. We use a depth 6 shallow layer Transformer to conduct this experiment. For the Transformer baseline model, we take the predictive entropy as uncertainty. For SNGP + Transformer, the entropy of the average of the Monte Carlo softmax samples is used as uncertainty. We do not compare with DUE for the CIFAR-100 dataset, as its training does not converge. SGD is used as the optimizer with the initial learning rate set to 0.01. All models are trained with batch size 128.

The accuracy, NLL, AUROC, and AUPR results are shown in Table \ref{tab:result_tiny}. The AUROC metric indicates the quality of uncertainty since it measures the probability that in-distribution (ID) and OOD samples can be separated \citep{Mukhoti2021DeterministicNN}. From the results, we have the following observations: 

(1) For OOD detection, \textit{The proposed LRFormer model outperforms all other methods with Transformer backbone on both CIFAR-10 vs SVHN and CIFAR-100 vs SVHN benchmarks}. This superior OOD detection performance benefits from the proposed LRSA regularization method, which solves both Lipschitzness and contraction problems in dot-product self-attention layers, and enables distance-preserving mapping in Transformer blocks.

(2) Notably, superior performance in OOD is achieved without sacrificing LRFormer's predictive performance. On the contrary, LRFormer even outperforms the standard Transformer baseline in terms of classification accuracy on the CIFAR-10 dataset, making LRFormer achieve the best performance in terms of all the metrics compared with all other single-network methods.

(3) Furthermore, the proposed LRSA self-attention can be computed efficiently using matrix operations, with minimal overhead compared to the original dot-product self-attention. This ensures LRFormer's performance gains come without compromising computation costs.

\subsubsection{Pre-training}
The recent work Plex~\citep{tran_plex_2022} comprehensively validated the reliability of the large pretrained models. The high performance of the Transformer results from pre-training on large-size datasets such as ImageNet-21K, and LAION-5B\citep{schuhmann2022laion}. It performs worse than CNNs if trained from scratch on small-size datasets. We use a depth 6 shallow layer Transformer to conduct pretrained weight experiments. The pretrained weights are loaded from the standard Transformer, sharing the same weight schemes of MLP layers and position embedding layer. This is because these layers in LRFormer have the same structure as the standard Transformer, so pre-trained weights can be directly applied to them. Our experiments in Table \ref{tab:pretrain_tiny} show that LRFormer can also benefit from these pre-trained weights. 

In summary, pre-trained weights of models can be directly transferred to LRFormer, which is very convenient for real-world applications.

\subsection{Visualization}
\label{sec:two_moons}
In order to show the interpretability of our model, we visualize the uncertainty heat map generated by LRFormer together with the baseline methods on the two moons 2D classification benchmark, which consists of two moon-shaped data clusters separable by a non-linear decision boundary. We employ a tiny Transformer architecture for this task, in which the depth is set to 9, the hidden dimension is set to 24 and the number of heads is set to 8. 

The uncertainty heat map comparisons are shown in Figure \ref{fig:two_moon}. Background color visualizes the predictive uncertainty of each model, where yellow stands for confidence and blue indicates uncertainty. All methods achieve 100\% test accuracy. From the results, we can observe that Deep Ensemble (Figure \ref{fig:ens}) estimates its uncertainty based on how far away test samples are from the decision boundary, without considering the data distribution. In Figure \ref{fig:due}, we can see that DUE without restrictions in the feature extractor, produces similar predictive uncertainty to Deep Ensemble which is heavily influenced by the distance from the decision boundary. Although SNGP can make allowances for data distribution, the decision boundary still has an impact on uncertainty estimation. The proposed LRFormer model, on the other hand, achieves near-ideal uncertainty quantification of this benchmark thanks to its bi-Lipschitz constraint in the LRSA self-attention layers, which allows it to maintain better distance awareness.

\section{Conclusion}

In this paper, we present LRSA, a regularization method designed to address overconfidence issues in Transformer structure models. By enforcing Bi-Lipschitz constraints and self-attention mapping contractions with theoretical guarantees, LRSA encourages the model to generate conservative predictions for out-of-distribution (OOD) inputs, thereby improving its ability to separate in-distribution (ID) data. 

Our approach primarily focuses on the attention mechanism of the Transformer architecture, which is a powerful and widely-used component in various natural language processing and vision tasks. Moving forward, it would be beneficial to extend our approach to incorporate these other modules within the Transformer architecture. Exploring how different combinations of modules can be leveraged to enhance performance across various tasks represents a promising avenue for future research. Additionally, investigating the relationship between Lipschitz regularity and other regularization techniques, including weight decay, dropout, and label smoothing, would provide valuable insights. Although these techniques have demonstrated effectiveness in preventing overfitting and improving generalization, their connection to Lipschitz regularity is not yet well-understood. Gaining a deeper understanding of this relationship could unlock insights into the inner workings of deep learning models and potentially lead to further performance improvements.

In conclusion, our proposed LRSA method addresses overconfidence issues in Transformer structure models by encouraging conservative predictions for OOD inputs. While our focus has been on the attention mechanism, future research directions involve incorporating other modules, exploring the relationship between Lipschitz regularity and other regularization techniques, and expanding LRFormer's applicability to diverse models and domains. These efforts contribute to advancing the field of deep learning and improving the robustness and performance of state-of-the-art models.

\begin{acknowledgements} 
    We thank all the anonymous reviewers for their insightful and thoughtful comments.
\end{acknowledgements}

\bibliography{ye_722,ma}

\onecolumn
\appendix

\title{Mitigating Transformer Overconfidence via Lipschitz Regularization\\(Supplementary Material)}

\maketitle
\section{Proof for the Lipschitz Constant of LayerNorm}

The LayerNorm operation~\citep{layernorm} used in LRFormer can be expressed as:
\begin{align*}
    \text{LN}(\mathbf{x}) &= \frac{\mathbf{x}-\mu(\mathbf{x})}{\sqrt{\sigma^2(\mathbf{x}) + \epsilon}} * \boldsymbol\gamma + \boldsymbol\beta 
\end{align*}
where $\mathbf{x}, \boldsymbol\beta, \boldsymbol\gamma \in \mathbb{R}^N$, $\mu(\mathbf{x}) = \frac{1}{N} \sum_{i=1}^N x_i$, $\sigma^2(\mathbf{x}) = \frac{1}{N}\sum_{i=1}^N (x_i - \mu(\mathbf{x}))^2$. 

WLOG, assume $N > 2$ and not all $x_i$ are equal.

The derivatives of $\mu$ and $\sigma^2$ w.r.t $x$:
$$\frac{\partial \mu}{\partial \mathbf{x}} = \frac{1}{N} \mathds{1}^\top$$ $$\frac{\partial \sigma^2}{\partial \mathbf{x}} = \frac{2}{N}(\mathbf{x} - \mu)^\top$$

Take the derivative of $\text{LN}(\mathbf{x})_i$, the $i$th element of $\text{LN}(\mathbf{x})$, with respect to $\mathbf{x}$ is:

\begin{align}
\begin{split}
    \frac{\partial \text{LN}(\mathbf{x})_i}{\partial \mathbf{x}}
    &= \gamma_i (\sigma^2 + \epsilon)^{-\frac{1}{2}} \bigg[(\mathbf{e}_i - \frac{1}{N}\mathds{1})^\top - \frac{1}{N} (\sigma^2 + \epsilon)^{-1} (x_i - \mu)(\mathbf{x} - \mu)^\top \bigg].
\end{split}
\end{align}

where $\mathbf{e}_I \in \mathbb{R}^N$ is a one-hot vector with $1$ at the $i$th element.
Therefore,
\begin{align*}
    \frac{\partial \text{LN}(\mathbf{x})}{\partial \mathbf{x}} &= (\sigma^2 + \epsilon)^{-\frac{1}{2}} \bigg[ \text{diag}(\boldsymbol\gamma) - \frac{1}{N}\boldsymbol\gamma \mathds{1}^\top - \frac{1}{N} (\sigma^2 + \epsilon)^{-1}\text{diag}(\boldsymbol\gamma)(\mathbf{x} - \mu)(\mathbf{x} - \mu)^\top \bigg].
\end{align*}

\begin{equation} \label{eq:first_terms_inf_norm}
    \left\Vert \text{diag}(\boldsymbol\gamma) - \frac{1}{N}\boldsymbol\gamma \mathds{1}^\top \right\Vert_{\infty} = \frac{2(N-1)}{N}\max_i |\gamma_i|,
\end{equation}

Take the infinity-norm on both sides, we have:
\begin{align*}
    \left\Vert \frac{\partial \text{LN}(\mathbf{x})}{\partial \mathbf{x}} \right\Vert_{\infty} &= (\sigma^2 + \epsilon)^{-\frac{1}{2}} \left\Vert   \text{diag}(\boldsymbol\gamma) - \frac{1}{N}\boldsymbol\gamma \mathds{1}^\top - \frac{1}{N} (\sigma^2 + \epsilon)^{-1}\text{diag}(\boldsymbol\gamma)(\mathbf{x} - \mu)(\mathbf{x} - \mu)^\top \right\Vert_{\infty} \\
    &\leq \epsilon^{-\frac{1}{2}} \bigg( \frac{2(N-1)}{N}\max_i |\gamma_i| + \frac{1}{N} \max_i |\gamma_i| N(N-2) \bigg) \\
    &\leq \epsilon^{-\frac{1}{2}} \max_i |\gamma_i| N.
\end{align*}

\section{Proof for the Lipschitz Constant of LRSA}
The pair-wise LRSA function is expressed as:
\begin{equation}
\label{LRSA}
    \begin{aligned}
    S_{ij} = -\frac{\alpha \left\Vert x_i^\top W_Q - x_j^\top W_K \right\Vert_2^2}{\left\Vert Q \right\Vert _F \left\Vert X^\top \right\Vert _{(\infty, 2)}}
\end{aligned}
\end{equation}

\begin{align*}
P_i = S_i(X)
\end{align*}

\begin{align*}
P_{ij} = \frac{e^{S_{ij}}}{\sum_{t=1}^n  e^{S_{it}} } \leq 1
\end{align*}

To take the derivative $P_{ij}$, there are two cases. 

When $t = j$:
\begin{equation}
    \begin{aligned}
\frac{\partial P_{ij}}{\partial S_{it}} &= \frac{\partial P_{ij}}{\partial S_{ij}} = 
\frac{\partial}{\partial S_{ij}}\bigg(\frac{e^{S_{ij}}}{\sum_{t=1}^n  e^{S_{it}}}\bigg) 
=
\frac{e^{S_{ij}}(\sum_{t=1}^n  e^{S_{it}}) - (e^{S_{ij}})^2 }{(\sum_{t=1}^n  e^{S_{it}})^2} \\ &= 
\frac{e^{S_{ij}}}{\sum_{t=1}^n  e^{S_{it}}}\bigg(1-\frac{e^{S_{ij}}}{\sum_{t=1}^n  e^{S_{it}}}\bigg) = P_{ij}(1- P_{ij})
\end{aligned}
\end{equation}

When $t \neq j$:
\begin{align*}
    \frac{\partial P_{ij}}{\partial S_{it}} = 
\frac{\partial}{\partial S_{it}}\bigg(\frac{e^{S_{ij}}}{\sum_{t=1}^n  e^{S_{it}}}\bigg) =
-\frac{e^{S_{ij}}}{\sum_{t=1}^n  e^{S_{it}}}\frac{e^{S_{it}}}{\sum_{t=1}^n  e^{S_{it}}} = -P_{ij}P_{it}
\end{align*}

\begin{equation}
    \begin{aligned}
    \frac{\partial P_{ij}}{\partial x_k}
    = \sum_{t=1}^n \frac{\partial P_{ij}}{\partial S_{it}} \frac{\partial S_{it}}{\partial x_k}
    = P_{ij}(1-P_{ij})\frac{\partial S_{ij}}{\partial x_k} - \sum_{t = 1, t \neq j}^nP_{ij}P_{it} \frac{\partial S_{it}}{\partial x_k}
    = P_{ij}\frac{\partial S_{ij}}{\partial x_k} - P_{ij}\sum_{t = 1} ^ n P_{it} \frac{\partial S_{it}}{\partial x_k}
\end{aligned}
\end{equation}

Take the infinity-norm on $S_{it}$, we get:
\begin{align*}
\left\Vert \frac{\partial S_{it}}{\partial x_k} \right\Vert _\infty & = \left\Vert \frac{\partial} {\partial x_k} \bigg( -\frac{\alpha \left\Vert x_i^\top W_Q - x_j^\top W_K \right\Vert_2^2}{\left\Vert Q \right\Vert _F \left\Vert X^\top \right\Vert _{(\infty, 2)}} \bigg) \right\Vert _\infty \\
& = \left\Vert -\frac{2 \alpha \left\Vert x_i^\top W_Q - x_j^\top W_K \right\Vert_2}{\left\Vert Q \right\Vert _F \left\Vert X^\top \right\Vert _{(\infty, 2)}}
\frac{\partial \left\Vert x_i^\top W_Q - x_j^\top W_K \right\Vert_2}{\partial x_k} + \frac{\alpha \left\Vert x_i^\top W_Q - x_j^\top W_K \right\Vert_2^2}{\left\Vert Q \right\Vert _F \left\Vert X^\top \right\Vert _{(\infty, 2)}^2}\frac{\partial \left\Vert X^\top \right\Vert _{(\infty, 2)}}{\partial x_k} \right\Vert _\infty \\
&\leq\left\Vert  \frac{2 \alpha \left\Vert x_i^\top W_Q - x_j^\top W_K \right\Vert_2}{\left\Vert Q \right\Vert _F \left\Vert X^\top \right\Vert _{(\infty, 2)}}
\frac{\partial \left\Vert x_i^\top W_Q - x_j^\top W_K \right\Vert_2}{\partial x_k}  \right\Vert _\infty
+ \left\Vert \frac{\alpha \left\Vert x_i^\top W_Q - x_j^\top W_K \right\Vert_2^2}{\left\Vert Q \right\Vert _F \left\Vert X^\top \right\Vert _{(\infty, 2)}^2}\frac{\partial \left\Vert X^\top \right\Vert _{(\infty, 2)}}{\partial x_k} \right\Vert _\infty \\
&\leq \frac{2\alpha}{\left\Vert Q \right\Vert _F} \frac{\left\Vert x_i^\top W_Q \right\Vert _2 + \left\Vert x_j^\top W_K \right\Vert_2}{\left\Vert X^\top \right\Vert _{(\infty, 2)}} 
\bigg( \frac{\partial \left\Vert x_j^\top W_Q \right\Vert_2}{\partial x_k} + \frac{\partial \left\Vert x_j^\top W_K \right\Vert_2}{\partial x_k} \bigg) + \frac{\alpha}{\left\Vert Q \right\Vert _F} 
\bigg( \frac{\left\Vert x_i^\top W_Q \right\Vert _2 + \left\Vert x_j^\top W_K \right\Vert_2}{\left\Vert X^\top \right\Vert _{(\infty, 2)}} \bigg) ^2 \\
&\leq \frac {2 \alpha (\left\Vert W_Q\right\Vert_2 + \left\Vert W_K \right\Vert_2) }{\left\Vert Q \right\Vert _F}^2 + \frac {\alpha (\left\Vert W_Q\right\Vert_2 + \left\Vert W_K \right\Vert_2) }{\left\Vert Q \right\Vert _F}^2 \\
& = \frac {3 \alpha (\left\Vert W_Q\right\Vert_2 + \left\Vert W_K \right\Vert_2) }{\left\Vert Q \right\Vert _F}^2
\end{align*}

Thus,
\begin{align*}
\left\Vert \frac{\partial P_{ij}}{\partial x_k} \right\Vert _\infty &= 
\left\Vert P_{ij}\frac{\partial S_{ij}}{\partial x_k} - P_{ij} \sum_{t = 1} ^ n P_{it} \frac{\partial S_{it}}{\partial x_k} \right\Vert _\infty \leq P_{ij} \frac {3 \alpha (\left\Vert W_Q\right\Vert_2 + \left\Vert W_K \right\Vert_2) }{\left\Vert Q \right\Vert _F}^2 + P_{ij}\sum_{t=1}^n P_{it} \frac {3 \alpha (\left\Vert W_Q\right\Vert_2 + \left\Vert W_K \right\Vert_2) }{\left\Vert Q \right\Vert _F}^2 \\
&\leq \frac {6 \alpha (\left\Vert W_Q\right\Vert_2 + \left\Vert W_K \right\Vert_2) }{\left\Vert Q \right\Vert _F}^2 \leq \frac{6 \alpha}{\left\Vert X \right\Vert _F} \cdot \frac { (\left\Vert W_Q\right\Vert_2 + \left\Vert W_K \right\Vert_2) }{\left\Vert W_Q \right\Vert _F }^2 
\end{align*}

% \begin{equation}
% S_{ij} = -\frac{\alpha}{\left\Vert Q \right\Vert ^F \left\Vert X^\top \right\Vert _{(\infty, 2)}}
% (x_i^\top W_Q - x_j^\top W_K)(x_i^\top W_Q - x_j^\top W_K)^\top
% \end{equation}
\section{Gaussian Process Layer}

As an optional module in LRFormer, Gaussian Process (GP) with an RBF kernel following SNGP \citep{Liu2020SimpleAP} is capable of perserving the distance awareness between input test sample and previously seen training data. This approach makes sure the model returns a uniform distribution over output labels when the input sample is OOD.

To make it end-to-end trainable, the Gaussian Process layer can be implemented a two-layer network:
\begin{equation}
    \operatorname{logits}(x)=\Phi(x) \beta, \quad \Phi(x)=\sqrt{\frac{2}{M}} * \cos (W x+b)
\end{equation}

Here, $x$ is the input, and $W$ and $b$ are frozen weights initialized randomly from Gaussian and uniform distributions, respectively. $\Phi(x)$ is Random Fourier Features (RFF) \citep{williams2006gaussian}. $\beta$ is the learnable kernel weight similar to that of a Dense layer. The layer outputs the class prediction $\operatorname{logits}(x) \in \mathbb{R}_{\operatorname{Num Classes}}$ .

\section{Experimental Details}

In Table~\ref{tab:supp-exp-details}, we provide the training details used for reproducing the main results in Tables above. The $Depth=12$ (pretraining) is the experimental setup of the ImageNet1K dataset pretraining. The other hyperparameters follows the same setting from DeiT III~\citep{Touvron2022ThreeTE}.

\begin{table}[!h]
    \centering
    \caption{Hyperparameters for LRFormer Training.} \label{tab:supp-exp-details}
    \begin{tabular}{lrrr}
      \toprule % from booktabs package
      \bfseries Hyperparameters & \bfseries $Depth=6$ & \bfseries $Depth=12$ & {\bfseries $Depth=12$} (pretraining) \\
      \midrule % from booktabs package
      Layer depth & 6 & 12 & 12 \\
      Input size & $224\times 224$ & $224\times 224$ & $224\times 224$ \\
      Batch size & 128 & 32 & 32 \\
      Warm-up steps & 5 & 5 & 5 \\
      Optimizer & SGD & AdamW & AdamW \\
      Learning rate & 0.01 & 0.006 & 0.004 \\
      Weight decay & 0.05 & 0.05 & 0.05 \\
      Learning rate scheduler & cosine & cosine & cosine \\
      Training epochs & 100 & 100 & 100 \\
      \bottomrule % from booktabs package
    \end{tabular}
\end{table}

\end{document}